# Feasible Static Workspace Optimization of Tendon Driven Continuum Robot based on Euclidean norm

Mohammad Jabari[1], Carmen Visconte[1] ,Giuseppe Quaglia[1], Med Amine Laribi[2]

[1] Department of Mechanical and Aerospace Engineering, Politecnico di Torino, Corso Duca degli Abruzzi 24, 10129 Torino, Italy
[2] Department of GMSC, Pprime Institute, University of Poitiers, CNRS, ISEA-ENSMA, UPR 3346, Poitiers, France
`{mohammad.jabari,carmen.visconte,giuseppe.quaglia}@polito.it`
`med.amine.laribi@univ-poitiers.fr`

**Abstract.** This paper focuses on the optimal design of a tendon-driven continuum robot (TDCR) based on its feasible static workspace (FSW). The TDCR under consideration is a two-segment robot driven by eight tendons, with four tendon actuators per segment. Tendon forces are treated as design variables, while the feasible static workspace (FSW) serves as the optimization objective. To determine the robot's feasible static workspace, a genetic algorithm optimization approach is employed to maximize a Euclidian norm of the TDCR's tip position over the workspace. During the simulations, the robot is subjected to external loads, including torques and forces. The results demonstrate the effectiveness of the proposed method in identifying optimal tendon forces to maximize the feasible static workspace, even under the influence of external forces and torques.

## 1 Introduction

Continuum robots, inspired by biological structures such as fish and snakes, offer exceptional flexibility and maneuverability, making them ideal for navigating confined spaces and interacting with unstructured environments [1,2]. These robots, characterized by their continuous tangent vector, can bend, stretch, contract, and twist along their length, enabling smooth motion in complex environments [3,4].

Typically, continuum robots feature a slender structure with a length significantly exceeding their other dimensions, a design particularly beneficial for medical applications. They are commonly a few millimeters in diameter and 20 to 30 centimeters long [5]. Their ability to minimize invasive procedures, improve patient comfort, and reduce recovery times has driven significant technological advancements in the medical field [6]. Traditional rigid-linked robots have limited flexibility, restricting access to delicate internal organs. Continuum robots address this limitation by enabling access to areas unreachable by conventional surgical tools [3,6]. Over the years, various continuum robot designs have emerged, enhancing the capabilities of minimally invasive surgical techniques [7]. The unique properties of continuum robots necessitate specialized design methodologies. Among these, the Newton-Euler method is widely utilized and has been extensively validated against other approaches, such as the finite element method (FEM) [8]. This method incorporates gravity effects on elastic tubes and vertebrae, providing reliable static analysis.

Continuum robots play a crucial role in human-robot interactive systems, where commands for desired movements are relayed through intuitive hardware interfaces. A proposed control system uses image feedback to manage both the robot's end-effector position and its configuration, ensuring conformity to desired paths [9]. Their adaptability and dexterity make continuum robots highly suitable for minimally invasive surgeries. Key attributes, such as a large workspace, high dexterity, and adequate payload capacity, have been the focus of numerous optimization studies [10]. For instance, research on cable-driven designs has explored the impact of segment numbers on performance in confined environments, revealing how friction influences robot deflection and how shape depends on tendon tension history [10]. Despite their flexibility, continuum robot design presents challenge due to the complexity of their kinematics, which involve curvature calculations [11]. Efficient computational methods are required to analyze dexterity while adhering to constraints [12]. Forward and inverse kinematics, crucial for configuration determination, depend on tendon force analysis. Mechanics-driven kinematic models have proven effective in addressing these challenges through optimization [13]. Energy efficiency is another critical aspect in tendon-driven continuum robots. Inspired by snake-like swimming motions, kineto-static analyses incorporate friction effects between cables and vertebrae, making energy consumption a key evaluation metric [1]. Continuum manipulators (CMs) are increasingly adopted in medical and industrial applications due to their adaptability and redundancy. Multi-segment structures with complex kinematics and larger workspaces offer improved performance but pose computational challenges, which remain an area of active research.

This paper introduces an optimal design for a two-segment tendon-driven continuum robot (TDCR) with eight tendons, focusing on maximizing workspace under external interactions. The kineto-static method is employed to determine the relationship between tendon force distribution and workspace. A genetic algorithm (GA) optimizes tendon tensions to achieve an optimal workspace configuration. This work employs static modeling to optimize FSW under external loads, prioritizing design simplicity over dynamic modeling's complexity, which is better suited for real-time motion analysis. Forward and inverse kinematic calculations are used to analyze workspace boundaries and robot configurations. Theoretical analysis and simulations validate the design's capability to achieve maximum workspace under environmental interactions.

The remainder of the paper is structured as follows: Section 2 details the static modeling of the proposed TDCR, incorporating backbone bending, torsion, friction, and environmental interactions. Section 3 explores feasible static workspace through optimization. Section 4 presents and discusses simulation results, and Section 5 concludes the study.

## 2 Robot modeling

The continuum robot in this study is a two-segment tendon-driven continuum robot (TDCR), featuring a compliant backbone with ten evenly spaced vertebrae per segment. Each segment is actuated by four symmetrically arranged tendons routed through the vertebrae for precise positioning and actuation Fig. 1b. The tendons are equally spaced around the backbone at 45° intervals and maintain an equal distance from the vertebra center Fig. 1d. Tendons within each segment are positioned at 90° intervals, ensuring balanced actuation and stability. When tension is applied, the tendons produce controlled bending, resulting in segmental curvatures $\theta_1$ and $\theta_2$ Fig. 1a. The tendons are actuated antagonistically for precise motion control.

The deformation of TDCR is conventionally modelled using the piecewise constant curvature (PCC) theory [1], which approximates the robot's shape as a series of tangent curved arcs, despite actual curvatures being non-circular. This model does not account for interactions with the external environment. Unlike kinematic modelling, static modelling considers tendon forces, friction, gravity, and external forces to analyse the robot's deformation, offering a more comprehensive understanding of its behaviour.

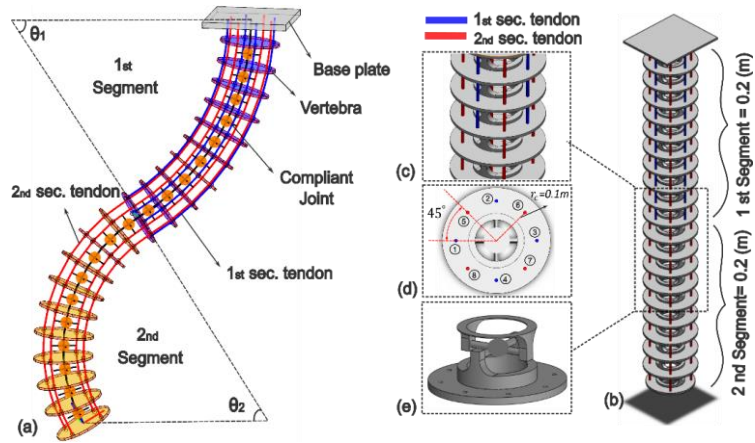

**Fig. 1.** (a) MATLAB model of the TDCR showing two groups of driven tendons and the bending angles $\theta_1$ and $\theta_2$ (b) CAD schematic of a two-segment, TDCR with four tendons in each segment (c) Anchor points for the tendons in Segment 1 (d) Distribution of tendons on the surface of the vertebra (e) Compliant joint.

### 2.1 Kinematic modeling of TDCR

In the kinematic modeling of the TDCR the constant curvature assumption is used. This means that the robot's vertebrae bend locally with a constant curvature [1]. While Cosserat formulations offer greater accuracy by modeling continuous deformations [14], our PCC approach simplifies the kinematics for appropriate offline workspace optimization, as detailed in Section 3. The conceptual picture of part of constant curvature TDCR is depicted in Fig. 2. Fig. 2a shows an image of consecutive vertebrae with two sample tendons and the notation and local coordinates attached to each vertebra along with the forces acting on a section of TDCR consisting of the backbone and vertebrae. Fig. 2b shows a magnified view showing the details of a universal joint with a backbone connected to each vertebra and the notation of the curvature angles describe the configuration of the robot components. Figs. 2. (c), (d), and (e) illustrate the internal forces and tendon forces acting on disk $i$ from disk $i+1$ to disk $i-1$.

Position of $d(s_i)$ of any point on the arc expressed as a function of the base in the $i-1$ frame is given by [1,5]

$$d(s_i) = \begin{bmatrix} x_i \\ y_i \\ z_i \end{bmatrix} = \begin{bmatrix} r_i \cos(\phi_i)(1 - \cos(\theta(s_i))) \\ r_i \sin(\phi_i)(1 - \cos(\theta(s_i))) \\ r_i \sin(\theta_i) \end{bmatrix} \quad (1)$$

in which $r_i = \frac{1}{k_i}$ is the radius of curvature. The subsegment curvature $k_i$ is defined by (2) [1,5],

$$k(s_i) = \sqrt{\beta_i^2 + \Gamma_i^2} \quad (2)$$

**Fig. 2.** (a) Subsegment kinematic diagram (b) Continuum robot pattern (c) Inter-Vertebral actuating forces applied on $i^{th}$ disk (d) Actuating moment on $i^{th}$ disk by resultant of inter-vertebral actuating forces (e) Disk and backbone masses and gravity forces for $i^{th}$ vertebra.

where $\beta_i$ and $\Gamma_i$ are the curvature of backbone along $\vec{x}_{i-1}$ and $\vec{y}_{i-1}$ , and $\phi_i$ is defined as Eq. (3)[1]:

$$\phi_i = tan^{-1}(\Gamma_i, \beta_i) \quad (3)$$

Given $d_i$ for $s_i = l_i$ , the homogeneous transformation matrix from $i-1$ to $i$ is given by [1,12]

$$^{i-1}T_i = \begin{bmatrix} R_z(\phi_i)R_y(\theta_i)R_z(-\phi_i) & d_i \\ 0 & 1 \end{bmatrix} \quad (4)$$

The direct geometrical model links the actuator space, giving the pulley angles, to the configuration space and to the cartesian space at the robot tip [1].

## 2.2 Static modelling of TDCR

According to the constant curvature static modeling theory, for given kinematic model the robot bending and its tip position are predicted. The basic formulation of the theory is briefly described below.

The position of each tendon described in local frame w.r.t the center of the disk is defined as [1]

$$^i p_{ti,c} = r_{c,m}[\cos(\theta_{m,c}) \quad \sin(\theta_{m,c}) \quad 0], m = 1,2, c = 1,2,3,4 \tag{5}$$

Where $\theta_{c,1} = \frac{2\pi}{4}(c-1)$ and $\theta_{c,2} = \theta_{c,1} + \frac{\pi}{4}$ . Here, $m$ and $c$ denote the segment index and tendon index, respectively.

The forces acting on the continuum robot under the quasi-static assumption consist of gravity, friction, and external forces. The gravity vector, defined in the global frame due to the mass of each disk and backbone, is transformed to the local frame. The gravity force acting on the disk is:

$$^{i-1}F_{d,i} = {}^{i-1}T_0 \, {}^0F_{d,i} \tag{6}$$

in which the gravity force $^0F_{d,i}$ is given as:

$$^0F_{d,i} = [0 \quad 0 \quad -m_{d,i} \cdot g \quad 0]^T \tag{7}$$

where $m_{d,i}$ is the mass of ith disk, and $^{i-1}T_0$ is the transformation of inertia to local frame.

The gravity force for the part of backbone is given as:

$$^{i-1}F_{b,i} = {}^{i-1}T_0 \, {}^0F_{b,i} \tag{8}$$

in which the gravity force $^0F_{b,i}$ is given as:

$$^0F_{b,i} = [0 \quad 0 \quad -m_{b,i} \cdot g \quad 0]^T \tag{9}$$

In which $m_{b,i}$ is the mass of $(i)^{th}$ part of backbone and $g$ is the acceleration of gravity. The resulted moment of gravity forces on the disks and part of backbone are described as [1]:

$$^iM_{d,i} = \left(\overrightarrow{^{i-1}o_{i-1} \, ^{i-1}o_i}\right) \times \left(^{i-1}F_{d,i}\right) \tag{10}$$

$$^iM_{b,i} = \left(\overrightarrow{^{i-1}o_{i-1} \, ^{i-1}p_{b,i-1}}\right) \times \left(^{i-1}F_{b,i}\right) \tag{11}$$

in which $^{i-1}o_{i-1}$ , $^{i-1}o_i$ and $^{i-1}p_{b,i-1}$ represent the coordinates of the center of $(i-1)^{th}$ and $(i)^{th}$ disks, and the center of gravity of the backbone between $(i-1)^{th}$ and $(i)^{th}$ disks in $(i-1)$ local frame, respectively.

In the case of interaction between TDCR and the surrounding environment, the external force $F_{ext}$ and torque $M_{ext}$ applied to the robot tip, denoted by index $v$ , and expressed in the tip coordinates, are described by the following relations [1]:

$$^{v-1}F_{ext} = {}^vT_0 \, {}^0F_{ext} \tag{12}$$

$$^{v-1}M_{ext} = \left(\overrightarrow{^{v-1}o_{v-1} \, ^{v-1}o_v}\right) \times \left(^{v-1}F_{ext}\right) \tag{13}$$

where $^0F_{ext}$ is the force applied to the tip of robot in inertial frame.

The driving tendon tension $^{i-1}F_{i,c}$ is the resultant of the two force vectors $F_{i,c}$ and $F_{i+1,c}$ by tendon $c$ illustrated in Fig. 2b, which are expressed using their unit direction vectors as bellow [1]:

$$^{i-1}F_{i,c} = F_{i,c}\left(\frac{\overrightarrow{^{i-1}p_{i,c}\,^{i-1}p_{i-1,c}}}{\left\|^{i-1}p_{i,c}\,^{i-1}p_{i-1,c}\right\|}\right) + F_{i+1,c}\left(\frac{\overrightarrow{^{i-1}p_{i,c}\,^{i-1}p_{i+1,c}}}{\left\|^{i-1}p_{i,c}\,^{i-1}p_{i+1,c}\right\|}\right) \tag{14}$$

$$^{i-1}M_{i,c} = {^{i-1}p_{i,c}} \times \left({^{i-1}F_{i,c}}\right) \tag{15}$$

The movement of the robot along the path occurs due to the forces applied by the tendons. As the configuration of robot changes, the frictional force is applied by tendons to the discs due to friction occurred at the contact point of tendons with the associated disk. The frictional force applied by each tendon to the disk is computed using the following equation.

The coefficient of friction is given as [1]

$$\mu_{i,c} = 0.689 e^{-0.027\sigma_{i,c}} \tag{16}$$

in which $\sigma_{i,c}$ is the local bending angle of the vertebrae determined using directional unit vectors as below:

$$^{i-1}U_{i-1,c} = \frac{\overrightarrow{^{i-1}p_{i,c}\,^{i-1}p_{i-1,c}}}{\left\|^{i-1}p_{i,c}\,^{i-1}p_{i-1,c}\right\|} \tag{17}$$

$$^{i-1}U_{i+1,c} = \frac{\overrightarrow{^{i-1}p_{i,c}\,^{i-1}p_{i+1,c}}}{\left\|^{i-1}p_{i,c}\,^{i-1}p_{i+1,c}\right\|} \tag{18}$$

$$\sigma_{i,c} = \cos^{-1}\left(\frac{^{i-1}U_{i-1,c} \cdot {^{i-1}U_{i+1,c}}}{\left\|^{i-1}U_{i-1,c}\right\|\left\|^{i-1}U_{i+1,c}\right\|}\right) \tag{19}$$

The tension magnitude $F_{i+1_c}$ is updated applying friction force between the disks and the tendons.

$$F_{i+1_c} = F_{i_c} - \mu_{i,c}\left|^{i-1}N_{i,c}\right| \tag{20}$$

$^{i-1}N_{i,c}$ is the normal force applied by tendon tension on the associated disk plane, which is computed using following relation [1]:

$$^{i-1}N_{i,c} = {^{i-1}F_{i,c}} - \left(^{i-1}F_{i,c} \cdot {^{i-1}n_i}\right) \tag{21}$$

Where $^{i-1}n_i$ is the unit normal vector of plane $(O_i, x_i, y_i)$.

The robot's backbone, as a support for the continuum robot structure, is considered as an elastic rod capable of bending in sides and twisting. Considering $K$ as the lateral bending stiffness coefficients and $E$ and $G$ as the elasticity and shear modulus of the TDCR supporting rod, the lateral static bending moment $M_b$ and torsion $M_\tau$ on each disk of the robot are calculated according to the following relations [5]:

$$^{i-1}M_{b,i} = {^{i-1}T_i}[0 \quad K \cdot E \cdot I \quad 0 \quad 0]^T \tag{22}$$

$$^{i-1}M_{\tau,i} = {^{i-1}T_i}[0 \quad 0 \quad 2G \cdot I \cdot (\varepsilon_i/l_i) \quad 0]^T \tag{23}$$

Where $\varepsilon_i$ is the twist of $i$th vertebra about z axis and $l_i$ and $I$ are the length and axial moment of inertia of corresponding backbone.

## 2.3 Equilibriums of the TDCR

The equilibrium relations of the TDCR include the recursive Newton-Euler equilibrium formulas for the forces and moments acting on each vertebra of the robot in its local coordinates, which are expressed as follows [1,5]:

$$i^{-1}F_{O_{i-1}} = \sum_{i}^{m} {}^{i-1}F_i + {}^{i-1}F_{d,i} + {}^{i-1}F_{b,i} + {}^{i-1}F_{O_i} \tag{24}$$

$$i^{-1}M_{O_{i-1}} = \sum_{i}^{m} {}^{i-1}M_{i,c} + {}^{i-1}M_{d,i} + {}^{i-1}M_{b,i} + {}^{i-1}M_{O_i} \tag{25}$$

The discretized forces and moments ${}^{i-1}F_{O_{i-1}}$ and ${}^{i-1}M_{O_{i-1}}$ are the lumped forces and moments at point $O_i$ in the frame $(i-1)$ [1].

Eq. (25) describes the elastic deformation and curvature of the central flexible backbone of the spatial planar TDCR using linear theory of tension with small strains in terms of the curvature of each section according to the relationship between curvature and bending moment. The equilibrium relation of the continuum robot is obtained from the equality of the net torque applied to the robot according to Eq. (26) and the torques resulting of the compliance property of bending and twisting of the robot from Eqs. (22), (23).

$$i^{-1}M_{O_{i-1}} = {}^{i-1}M_{b,i} + {}^{i-1}M_{\tau,i} \tag{26}$$

By solving the above nonlinear coupled equations, the static configuration of the TDCR is determined [1,4].

## 3    Feasible Static Workspace of TDCR

Tendon force-based workspace is a key performance metric for tendon-driven continuum robots (TDCRs), though research in this field remains in its early stages. A robot's workspace defines the region encompassing all reachable positions, primarily determined by its structural parameters. It is typically derived from the forward kinematics model, which considers the translational motion of the end effector. The workspace is commonly computed using the homogeneous transformation matrix, which establishes the relationship between the end-effector and base coordinate frames [10].

$$^{0}T_n = {}^{0}T_1 \, {}^{1}T_2 \cdots {}^{n-1}T_n \tag{27}$$

This function represents the configuration of the robot's tip coordinate system relative to the base coordinate system and is expressed in the following general form. [10,12]:

$$^{0}T_n = \begin{bmatrix} {}^{0}R_n & {}^{0}P_n \\ 0 & 1 \end{bmatrix} \tag{28}$$

in which the ${}^{0}R_n$ block corresponds to the rotation, and ${}^{0}P_n$ block indicates the position vector of the robot's tip from the base.

The feasible static workspace in this study is defined as the robot's maximum reach relative to its base, represented by the maximum length of the tip position vector ${}^{0}P_n$. This metric is used as the performance criterion for evaluating the robot, as detailed in the following section.

### 3.1 Objective function

This section presents the process of achieving a feasible static workspace (FSW) for a TDCR using a genetic algorithm (GA) optimization to determine the optimal tendon force values. The optimization problem, focused on maximizing the static workspace, is typically reformulated as minimizing the reciprocal of the Euclidean norm of the TDCR's tip position. It can be expressed as follows:

$$\underset{i=1,\cdots,N_t}{Minimize}\left[f\big(^{i-1}F_{i,c}\big) = \left\| \, ^0P_n\right\|_2^{-1}\right]$$
$$s.t. \ \Theta(\mathcal{F},\mathcal{S}) = 0 \tag{29}$$
$$F_{i,c}^l \leq F_{i,c} \leq F_{i,c}^u, i = 1,\cdots,N_t$$

Here, $F_{i,c}$ denotes the set of tendon forces used as design variables in the optimization task, constrained by the lower and upper bounds $F_{i,c}^l$ and $F_{i,c}^u$, respectively. $\Theta(\mathcal{F},\mathcal{S})$ represents the kineto-static model of the TDCR, as defined by Eq. (26). This model is a function of the tendon forces $\mathcal{F}$ and the robot's configuration parameters $\mathcal{S}$, which include the set of curvatures about the local x- and y-axes, and the twist about the z-axis.

A GA method optimization is employed to determine the FSW of the TDCR. During the optimization process, the robot is assumed to be subjected to a specified external force and moment. The population size was tested with values of 50, 100, and 200 chromosomes, and the convergence criterion was set to an accuracy of $10e^{-6}$. The results demonstrated that the desired accuracy could be achieved using the smallest population size of 50 individuals, requiring a maximum of five iterations, which imposed minimal computational cost.

## 4 Simulation and results

As previously mentioned, determining the static workspace serves as the objective function in the optimization problem addressed in this paper. A representation of this workspace, generated for 100 randomly selected tendon force values, is depicted in Fig. 3(a). The figure includes discrete data points corresponding to the randomly generated values, as well as a spline approximation representing the continuous curve. The vertical axis indicates the norm of the robot tip's position vector, $\left\| \, ^0P_n\right\|_2$, measured from the base, while the horizontal axis corresponds to the test number associated with each randomly selected set of tendon force vectors.

Fig. 3(a) displays two datasets: a scatter plot of discrete data points (blue circles) and a continuous red spline curve approximating the data. It highlights the robot's workspace complexity as a function of design variables and the challenge of identifying the optimal point or solving the feasible workspace problem by maximizing the curve. A simulation evaluated the robot's performance under external force (0.1, 0.1, -0.1) $N$ and torque (-0.1, 0.1, 0.1) $N \cdot m$ along the x, y, and z axes Fig. 3b. Fig. 4 shows the robot's configurations under 10 random tendon force samples, revealing the relationship between tendon forces and workspace mobility. A genetic algorithm is used to optimize the feasible static workspace (FSW) by selecting tendon forces as design variables.

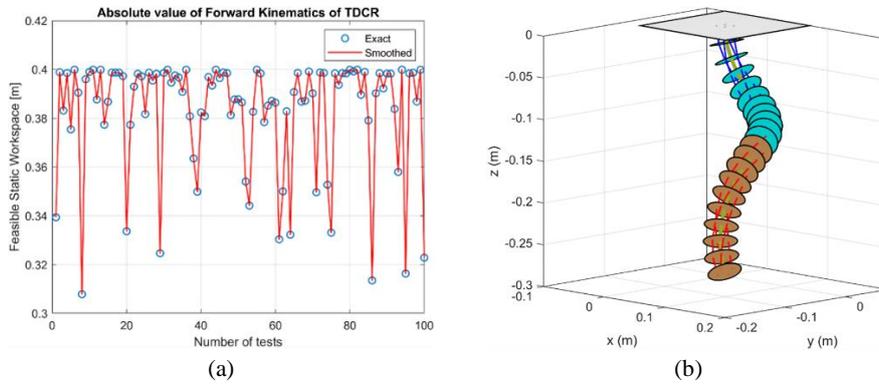

(a)                                                    (b)

**Fig. 3.** (a) Feasible Static Workspace for 100 randomly sample design variables. (b) A schematic representation of the robot's configuration under an applied external force and torque is shown.

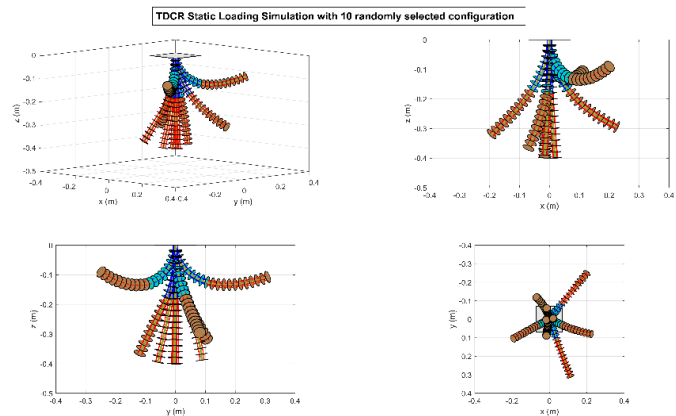

**Fig. 4.** Robot configuration in selecting ten random sample states of tendon forces.

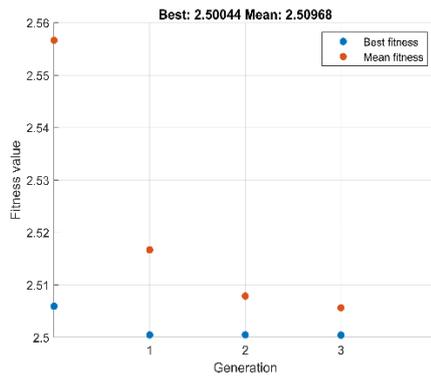

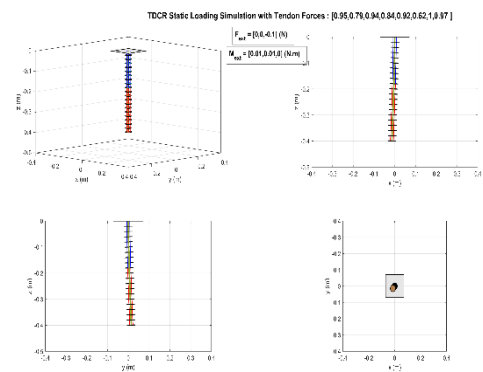

**Fig. 5.** The trend of objective function changes during optimization

**Fig. 6.** Optimal robot position in accessing the maximum possible static workspace in the presence of a vertical external force

Fig. 5 illustrates the progression of changes in the objective function. Based on the robot's structure detailed in Fig. 1, the maximum length of the robot, comprising two segments with 10 vertebrae per segment and a backbone spacing of 0.2 meters, is 0.4 meters, with its inverse being 0.25 meters. Referring to the graph in Fig. 5, the cost of the best sample in the population is represented by blue points, while the average population cost is shown by red points. The convergence trend progresses until the cost function reaches its minimum value, corresponding to the maximum static workspace, which represents the feasible workspace. From the observed trend, it is evident that optimization is achieved by the fourth generation, with the population average converging closely to the minimum value.

The optimal robot position for accessing the maximum FSW under a vertical external force, caused by the mass of an object attached to the end effector, is depicted in Fig. 6. As shown, the optimizer successfully calculates the optimal tendon forces, counteracting the effect of the external force, and achieves the maximum feasible static workspace, represented by the position illustrated in the figures.

## 5 Conclusion

This study proposes an analysis of the feasible static workspace (FSW) of a tendon-based continuum robot (TDCR) with two segments and four tendons per segment, utilizing genetic algorithm (GA) optimization. Tendon forces are treated as design variables, while the reciprocal of the Euclidean norm of the TDCR's tip position is used as the optimization objective to maximize the feasible static workspace. Simulation results highlight the effectiveness of the proposed approach in determining the optimal tendon forces to maximize the static workspace, even when the tendon-driven continuum robot (TDCR) is subjected to external forces and moments.

**Acknowledgments** This publication is part of the project PNRR-NGEU which has received funding from the MUR – DM 351/2022.

**Nomenclature**

| Symbol | Description | Symbol | Description |
|---|---|---|---|
| $m$ | Total number of sections | $^{0}F_{e}$ | External force applied at the tip of the robot in the global frame |
| $c$ | Designated tendon among $m$ sections | $^{i-1}F_{i,c}$ | Driving tendons tensions applied by tendon $c$ on disk $i$ in the frame $i-1$ |
| $i$ | Index of vertebra along the robot backbone | $^{i-1}M_{i,c}$ | Moment induced by tendon tension in the frame $i-1$ |
| $\theta$ | Bending angle | $^{i}p_{ti,c}$ | Tendon distribution array on disk $i$ |
| $\varphi_i$ | Angle defining the direction of curvature | $^{v-1}F_{ext}$ | External force applied at the tip of the robot in the frame $i-1$ |
| $r_i$ | Radius of curvature of the subsegment | $^{v-1}M_{ext}$ | Moment induced by External force in the frame $i-1$ |
| $k_i$ | Subsegment curvature | $m_{d,i}$ | Mass of the disk i |
| $\Gamma_i$ | Curvature of the backbone along $x_{i-1}$ axis | $m_{b,i}$ | Mass of the backbone between disk $i$ and $i-1$ |
| $\beta_i$ | Curvature of the backbone along $y_{i-1}$ axis | $^{i-1}U_{i-1,c}$ | Unit vector between the frames $i$ and $i-1$ in the frame $i-1$ |

| $d(s_i)$ | Position of any point along the arc in the $i\text{-}1$ frame | $^{i-1}U_{i+1,c}$ | Unit vector between the frames $i$ and $i + 1$ in the frame $i - 1$ |
|---|---|---|---|
| $^{i-1}T_i$ | Homogeneous transformation matrix from frame $i$ to $i\text{-}1$ | $^{i-1}N_{i,c}$ | Normal force in the frame $i - 1$ |
| $^{i-1}T_0$ | Homogeneous transformation matrix from frame $0$ to $i\text{-}1$ | $^{i-1}n_i$ | Unit normal vector in the frame $i - 1$ |
| $R\,z(\phi i)$ | Rotation matrix around $z_{i-1}$ axis | $\mu$ | Friction coefficient |
| $R\,y(\theta i)$ | Rotation matrix around $y_{i-1}$ axis | $\sigma$ | Local bending angle of the vertebrae |
| $R\,z(-\phi i)$ | Rotation matrix around $z_{i-1}$ axis | $N_i$ | Normal force applied by tendon tension |
| $^{i}p_{i,c}$ | Tendon position on disk $i$ | $E$ | Young's modulus of the backbone material |
| $F_{i+1,c}$ | Tension magnitude along the tendon $c$ between the frame $i$ and $i + 1$ | $G$ | Shear modulus of the backbone material |
| $\left\Vert\,^0P_n\right\Vert_2$ | Euclidean distance from the base of the robot to its tip | $I$ | Moment of inertia of the backbone cross-section |
| $^{i-1}F_{d,i}$ | Force due to gravity on disk $i$ | $M_b$ | Bending moment |
| $^{i-1}F_{b,i}$ | Force due to gravity on backbone between disk $i$ and $i\text{ -}1$ | $M_\tau$ | Torsional moment |
| $^{i}O_i$ | Center of disk $i$ in the frame $i$ | $FSW$ | Feasible static workspace |
| $r_c$ | Radius from center of disk $i$ | $F_{i,c}$ | Driving tendon force applied by tendon $c$ between the frame $i - 1$ and $i$ |
| $F_{ext}$ | External force applied at the robot tip | $M_{ext}$ | External torque applied at the robot tip |